\documentclass{article}

\usepackage[preprint]{neurips_2026}
\makeatletter
\renewcommand{\@noticestring}{Preprint. Under review.}
\makeatother

\usepackage[utf8]{inputenc}
\usepackage[T1]{fontenc}
\usepackage{hyperref}
\usepackage{url}
\usepackage{booktabs}
\usepackage{graphicx}
\usepackage{amsmath}
\usepackage{amsfonts}
\usepackage{nicefrac}
\usepackage{microtype}
\usepackage{xcolor}
\usepackage{colortbl}

\usepackage{textcomp}
\newcommand{\sq}{\textquotesingle}

\usepackage{amssymb}
\usepackage{listings}

\definecolor{PrefixAccepted}{RGB}{30,64,175}
\definecolor{SuffixAccepted}{RGB}{4,120,87}
\definecolor{RegularDecoded}{RGB}{90,90,90}
\definecolor{RejectedDraft}{RGB}{185,28,28}
\definecolor{SpeedupGreen}{RGB}{220,252,231}

\newcommand{\gaincell}[1]{\cellcolor{SpeedupGreen}{#1}}

\lstdefinestyle{selfspecexample}{
  basicstyle=\ttfamily\scriptsize,
  columns=fullflexible,
  keepspaces=true,
  breaklines=true,
  showstringspaces=false,
  escapeinside={(*@}{@*)},
  frame=none,
  numbers=none
}

\usepackage{algorithm}
\usepackage{algpseudocode}
\algrenewcommand\algorithmicreturn{\textbf{return}}

\algnewcommand{\Input}{\item[\textbf{Input:}]}
\algnewcommand{\Output}{\item[\textbf{Output:}]}

\title{Self-Speculation for Faster Reasoning Models}

\author{
  Ravisri Valluri\thanks{Correspondence to \texttt{ravisrivk@gmail.com}.} \\
  University of California, Los Angeles \\
  \texttt{ravisrivk@gmail.com} \\
  \And
  Tung Nguyen \\
  University of California, Los Angeles \\
  \And
  Aditya Grover \\
  University of California, Los Angeles \\
}

\begin{document}

\maketitle

\begin{abstract}
Large language models (LLMs) are deployed for increasingly complex tasks involving planning and multi-step decision making, but high-quality performance on these tasks often requires generating long reasoning traces. This is a poor fit for latency-sensitive and interactive applications like voice assistants or coding agents, where generation latency can strongly affect user experience. Existing acceleration methods typically focus on token-level generation, without utilizing the structure of reasoning workflows. We introduce SSR: Self-Speculation for Reasoning Models, a training-free self-speculative decoding method that leverages the chain-of-thought (CoT) as a source of speculation. SSR uses the partial-CoT answer distribution as the drafter and the full-CoT distribution as the verifier, deriving both from the same model at different reasoning budgets. This builds on the observation that later partial-CoT responses often exhibit greater semantic and lexical overlap with the full-budget response. Due to this overlap, SSR can accept long draft prefixes at once, leading to large speedups on structured and long-form generation tasks. To further exploit draft-response overlap beyond the contiguous prefix accepted by standard speculative decoding, SSR also incorporates suffix decoding, using the draft to seed a suffix cache and recover useful spans beyond the accepted prefix, further reducing latency on tasks with high lexical overlap between the draft and the final response. We evaluate SSR on multiple structured and long-form generation tasks where it is most useful, and demonstrate a relative improvement of up to 24.1\% on total generation latency for popular open-source models such as Qwen3.5 and Gemma-4.
Code is available at \url{https://github.com/Ravi-VK/SSR/}.

\end{abstract}

\section{Introduction}

Reasoning language models are pushing LLMs beyond short-form assistance and into tasks that require sustained problem solving, including planning, coding, and decision making. These models achieve stronger performance by generating long reasoning traces that can span over thousands of tokens before producing a final response. While this improves answer quality, it also increases end-to-end generation latency, making them difficult to use in applications that require fast responses. This is especially limiting in interactive settings such as chatbots, voice-based assistants, and coding agents, where a long delay can degrade the user experience. In such settings, it is often worth spending additional compute if doing so reduces end-to-end latency.

Speculative decoding is a natural fit for this setting.
In standard speculative decoding, a draft model proposes multiple future tokens, which are then verified in parallel by the target model \citep{specdecoding, specsampling}.
When the draft is accurate, the verifier can accept several tokens at once and reduce the number of sequential decoding steps.
The main challenge is constructing a draft that is both cheap to produce and accurate enough that the target accepts most of its tokens.

Prior work on speculative decoding has approached draft model selection in different ways. A fully independent draft model is the natural baseline, but it introduces substantial overhead, since the draft model must be trained separately, kept in memory alongside the target model, and carefully optimized to match the target model's output distribution. One line of work, including Medusa \citep{medusa} and EAGLE \citep{eagle}, reduces this overhead by training lightweight draft heads or small modules that reuse parts of the target model, but still requires training and architectural modifications. Another line of work avoids auxiliary parameters entirely by drafting from the target model itself, using early layers \citep{layerskip, EESD}, sparse attention \citep{specattn, sparsespec}, quantization \citep{quantspec}, or other approximations of the target model's internal computation. Quantization- and sparse-attention-based variants of self-speculative decoding can often be entirely training-free. At the other end of the spectrum, model-free methods \citep{promptlookupngram, suffixdecoding, samdecoding} copy text from the prompt or prior generations using n-gram lookup or suffix structures. These methods eliminate draft-model calls entirely, but cannot generate new content or adapt to the model's current reasoning state.

While all these approaches can be effective at reducing decoding latency, they do not fully take advantage of the structure of reasoning language models.
Reasoning language models expose an additional source of signal for speculation: the chain-of-thought itself.
Work on test-time scaling has shown that CoT length can be controlled through budget forcing and reflection \citep{s1}, and that increasing the reasoning budget generally improves answer quality, at least up to a point.
This suggests that partial CoTs are not completely arbitrary incomplete states, and that they can induce meaningful intermediate answer distributions.
Answers sampled early in the CoT might be incomplete or inconsistent, but as reasoning progresses they improve in quality and become increasingly similar in content and surface form to the final answer.
This property offers a natural basis for drafting, and prior work has not used it so far.

\begin{figure}[t]
    \centering
    \includegraphics[width=0.98\linewidth]{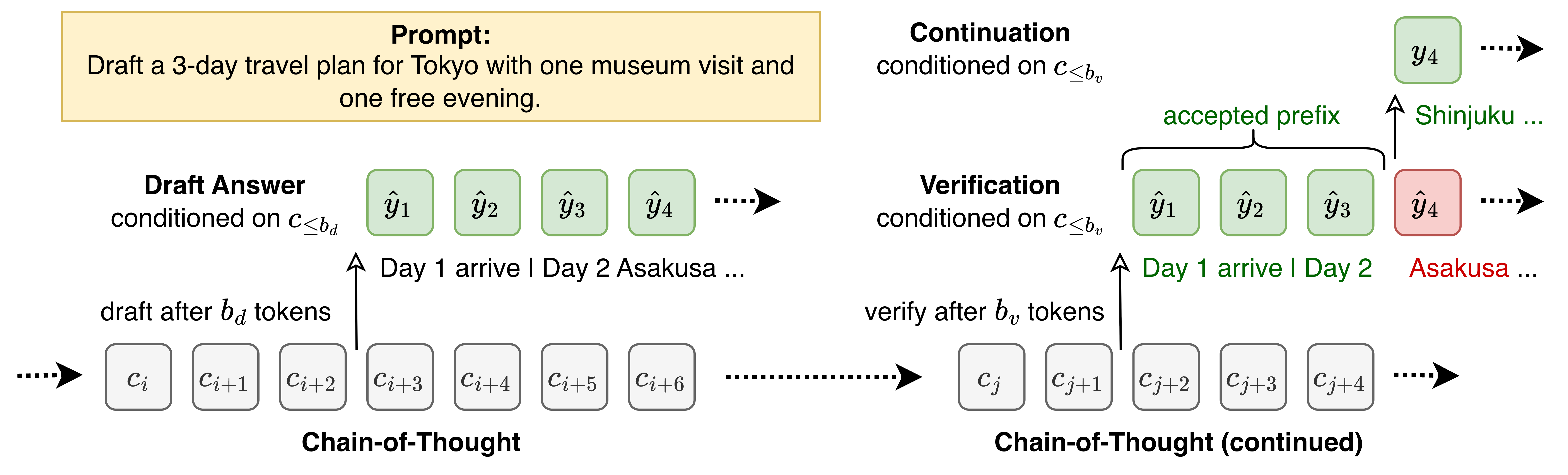}
    \caption{Illustration of concurrent generation in self-speculative decoding. A lower-budget instance drafts answer tokens while reasoning continues in parallel. Then the verifier accepts the longest draft prefix supported by the higher-budget distribution; after the first rejected token, generation continues under the verifier.}
    \label{fig:one_shot_self_spec}
\end{figure}

We take advantage of this property to propose SSR: Self-Speculation for
Reasoning Models, a self-speculative decoding method that uses the answer
distribution induced by a partial CoT as the draft distribution.
The same language model is used at two different reasoning budgets --- an
intermediate CoT prefix produces draft answer tokens, while the completed
CoT acts as the verifier. Draft generation and continued reasoning proceed
concurrently, so the drafting overhead is hidden behind ongoing CoT
generation (Figure~\ref{fig:one_shot_self_spec}).
Because the draft distribution converges toward the final answer distribution
as the CoT grows, SSR requires no auxiliary parameters or target-model-specific
modifications. Under exact verification, output quality is unchanged. When
drafts are accepted, latency decreases, and when rejected, the model falls
back to standard decoding with minor overhead, making SSR a drop-in
replacement for standard decoding in any reasoning model.

We perform left-to-right draft verification to accept the longest matching prefix between the draft and the final answer. In structured tasks, the model commits to broad structure early in the trace, so a draft sampled mid-CoT often shares substantial overlap with the final response. For long-form outputs, however, two responses might disagree on minor details like variable names or phrasing choices early on, while still agreeing on long spans later in the text. Discarding the draft at the first rejected token wastes this overlap. We therefore follow this prefix verification stage with suffix decoding, building a suffix cache from the draft and using it to propose spans that may still match beyond the first rejection, recovering useful text that standard verification would typically discard.

If the draft acceptance rate increases monotonically with CoT length, the optimal sampling point would leave exactly enough CoT tokens remaining to mask the drafting cost in order to begin sampling the draft as late as possible. In practice, this ideal point is not known in advance, since it depends on the total CoT length.
This motivates an iterative variant of SSR that samples drafts at multiple points during CoT generation. Earlier drafts are not wasted — we apply the same verification procedure to bootstrap later drafts from earlier ones, reducing the overhead of multi-point sampling and increasing the likelihood that at least one draft overlaps substantially with the final answer.

In summary, our contributions are:
\begin{itemize}
    \item We propose SSR, a training-free self-speculative decoding method for reasoning language models that uses partial-CoT answer distributions as drafts, requiring no auxiliary parameters or target-model-specific modifications.

    \item We extend standard draft verification with suffix decoding to recover useful draft spans beyond the first mismatch, improving acceptance rates for long-form and structured outputs.

    \item We introduce an iterative variant that samples drafts at multiple points during CoT generation, using earlier drafts to bootstrap later ones and reduce multi-point sampling overhead.

    \item We provide an implementation in vLLM that integrates SSR as a drop-in addition to existing serving configurations, requiring no changes to model weights or inference infrastructure.

    \item We evaluate on coding and long-form generation tasks including HumanEval, ClassEval, and LongProc, achieving latency reductions of up to 24.1\% while preserving output quality.
\end{itemize}

\section{Related Work}
\label{related_work}

\subsection{Speculative decoding}
The key driver of generation latency for autoregressive language models
is the decode step, which is repeated across a large number of tokens.
Speculative decoding reduces this cost by using a cheaper draft model
to propose multiple tokens in advance, which are then verified in
parallel by the target model. In its exact formulation, it uses
rejection sampling to ensure that the final sampling distribution
remains identical to that of the target model. If multiple draft tokens
are accepted, this can significantly reduce the number of expensive
decoding steps required of the target model. This line of work
\citep{specdecoding, specsampling} was developed as a lossless
acceleration, but has since expanded to include lossy variants,
including model-free speculative decoding, which does not necessarily
associate a sampling probability with draft tokens, and principled
relaxations of verification \citep{sprinter} that can achieve even
higher acceleration gains.

\paragraph{Draft model construction.} The speedup speculative decoding provides is a function of both
drafting overhead and acceptance rate. For speculative decoding to
accelerate generation, the draft model must be cheap to run and close
enough to the target model to maintain a high acceptance rate. In the
open-source community, it is common to use a smaller, distilled variant
of a flagship model as the draft model, especially when the target
reaches tens or hundreds of billions of parameters. But when resources
are constrained, and because the draft model also occupies significant
memory, many popular approaches share parameters and computation between
draft and target to minimize this overhead. \citet{eagle} predicts
penultimate-layer hidden states for future positions using lightweight
feature predictors, then uses the last layer of the target model to
predict the corresponding tokens in parallel as the draft proposal.
\citet{medusa} adds additional language modeling heads to decode future
positions from the same hidden state. These approaches substantially
reduce the footprint of drafting, but still require target-model-specific
training and architectural modifications.

\paragraph{Self-speculative decoding.} Another line of work eliminates the draft model entirely, using only the parameters and activations of the target model to propose tokens. Self-speculative decoding typically relies on a lossy approximation of the target model for draft proposals — via early exiting, sparsity, or quantization — and reruns full inference for verification. \citet{layerskip} and \citet{EESD} enable self-speculative decoding through early exiting, with the former training intermediate layers to draft tokens for full-model verification and the latter adding a Thompson-sampling-based controller to decide when to exit and speculate. While these layer-skipping approaches typically require training, quantization and sparsity-based approaches can be entirely training-free. \citet{quantspec} uses a hierarchical quantized KV cache to create a cheaper self-speculative draft pass while preserving the full-precision cache for verification. \citet{specattn} co-designs sparse attention with self-speculative decoding so that the draft pass reads only a subset of the KV cache before full verification. \citet{sparsespec} also uses sparse attention, proposing PillarAttn, which uses attention scores to dynamically identify a small subset of important tokens for long-reasoning workflows. None of these approaches exploits the structure of reasoning models directly: even PillarAttn, which targets long CoT generation, treats the chain of thought as a sequence to attend over sparsely rather than exploiting how reasoning converges to a final answer.

\paragraph{Model-free methods.} Self-speculative decoding approaches still incur overhead from a  relatively cheaper but nonetheless non-trivial drafting pass. At the extreme end of the spectrum, model-free approaches can propose spans of text with negligible overhead. N-gram speculative decoding \citep{promptlookupngram} proposes fixed-length n-grams drawn from the prompt and previously generated tokens, working well in domains with high lexical overlap. Suffix decoding and related work \citep{suffixdecoding, samdecoding} extend this by building a suffix tree over this text, enabling longer span proposals based on the currently matched suffix of the generated tokens. These methods are fast but inherently passive — they can only repeat text that already exists somewhere, and cannot adapt meaningfully to the model's current reasoning state.

\subsection{Inference-time compute and budgeted reasoning}
Recent work has shown a clear tradeoff between inference-time compute
and answer quality in reasoning language models, driving methods to
explicitly control reasoning length at test time. Budget forcing
controls the number of thinking tokens by forcing early termination
or extending reasoning — for example by appending continuation tokens
— with answer quality generally improving as the reasoning budget grows
\citep{s1}. A parallel line of work studies when model outputs
stabilize during a reasoning trace, using stability signals to stop
early and reduce redundant computation \citep{convergenceearlystop}.
More broadly, work on test-time compute scaling examines how to
allocate additional inference compute to improve answer quality
\citep{snellscale, welleckscale}. These methods primarily treat
intermediate-budget outputs as cheaper final predictions or early-stop
endpoints, rather than as computation that can be reused for subsequent generation.

\section{Preliminaries}
\label{preliminaries}
\subsection{Autoregressive reasoning language models}
We focus on applications where a language model generates a reasoning trace before producing a final response. The chain-of-thought is sampled first, token by token, until an end-of-thinking token is reached. The final response is then conditioned on both the user query and the reasoning trace, and is sampled token by token until an end-of-sequence token is reached. In budget forcing, the language model is used to sample a response after a specified number of CoT tokens; sampling a response after partial CoT typically requires appending a special end-of-thinking token and, in some setups, an interrupt string.

We use $p_\theta(o \mid q)$ to denote the probability of sampling output sequence $o$ for user input $q$ from a language model with parameters $\theta$. For autoregressive models, $ p_\theta(o \mid q) = \prod_{t=1}^{l} p_\theta(o_t \mid x, o_{<t}) $, where $o = (o_1, \ldots, o_l)$. We denote the reasoning trace (chain-of-thought) by $r$ and the final answer sequence by $a$: for reasoning language models, the output sequence is the concatenation of the reasoning trace and the final answer. If a reasoning budget $b$ is specified, we write $p_\theta(a \mid q, b)$ as shorthand for $p_\theta(a \mid q, r_{\leq b})$, where $r_{\leq b}$ denotes the first $b$ tokens of the reasoning trace.

\subsection{Speculative decoding}
In speculative decoding, a draft model with distribution $d$ proposes  tokens, which are then verified by the target model with distribution  $p$. The draft model proposes a sequence of $n$ tokens $\hat{a}_1, \ldots, \hat{a}_n$ before verification begins. Tokens are accepted sequentially until the first rejection; we refer to the accepted tokens as the accepted prefix. A draft token $\hat{a}_t$ is accepted with probability

\[
\alpha_t = \min\left(1, \frac{p(\hat{a}_t \mid q, r, a_{<t})}{d(\hat{a}_t \mid q, r, a_{<t})}\right),
\]

which ensures that accepted tokens are distributed according to $p$.
Upon rejection, a token is sampled from the residual distribution

\[
a_t \sim \frac{\max(p - d, 0)}{\sum_v \max(p(v) - d(v), 0)},
\]

where the residual is the positive part of $p - d$, renormalized to a
valid distribution. Together, acceptance and residual sampling guarantee
that the final output distribution is identical to that of the target
model $p$, regardless of the draft model quality.

The draft model proposes a sequence of tokens before verification, so
if the acceptance rate is high, many tokens are accepted in a single
verifier forward pass, yielding a net speedup. Exact speculative
decoding requires maintaining both $p$ and $d$ at each position to
compute the residual upon rejection; simply resampling from $p$ upon
rejection yields an approximate variant that is also widely used in
practice \citep{li2025nvidia_specdec_intro}.

\subsection{Suffix decoding}
Suffix decoding \citep{suffixdecoding} extends speculative decoding to
the model-free setting, where no draft distribution $d$ is available.
Rather than sampling draft tokens from a model, it indexes all suffixes
of the prompt and previously generated tokens into a suffix cache, and
looks up continuations matching the current generation context as draft
proposals. Since no draft probabilities are associated with proposed
tokens, exact rejection sampling is not applicable. Instead, proposed
tokens are verified greedily and accepted only if they match the
target model's greedy prediction, making suffix decoding an approximate
rather than exact acceleration method. It works best in settings with
high lexical repetition, such as agentic loops and code editing, where
matching suffixes are likely to reoccur.

\section{Method}
\label{method}

\subsection{Self-Speculative Decoding}
We instantiate speculative decoding using a single reasoning model
evaluated at two CoT budgets, with draft distribution
$p_\theta(a \mid q, b_d)$ and verifier distribution
$p_\theta(a \mid q, b_v)$, where $b_d < b_v$. In practice,
conditioning on a partial CoT $r_{\leq b}$ requires appending an
end-of-thinking token and, in some setups, a budget-forcing prefix
to prompt the model to produce an answer. The key observation is
that draft generation and continued CoT reasoning can proceed
concurrently: while the model continues reasoning toward budget
$b_v$, it simultaneously produces draft answer tokens conditioned
on $r_{\leq b_d}$. The overhead of drafting is therefore largely
hidden behind ongoing reasoning.

For concurrent drafting to be feasible, the total CoT length must
exceed the draft budget plus the maximum draft length,
\[
|r| > b_d + m,
\]
so that draft generation completes before CoT generation ends.
Beyond this prerequisite, draft quality depends on how much
reasoning has completed by budget $b_d$: when $b_d$ is a small
fraction of the total CoT length, the partial CoT may not yet
have converged toward the final answer, producing drafts that
diverge in high-entropy tasks. In practice, we find that $b_d$
should be a substantial fraction of the total CoT length for
drafts to be useful.

At a high level, the procedure has four phases: (i) generate
reasoning tokens until the draft budget $b_d$ is reached, (ii)
concurrently sample a draft answer $\hat{a}$ conditioned on the partial
CoT $r_{\leq b_d}$ while reasoning continues, (iii) collect
the draft once reasoning reaches $b_v$, and (iv) verify the
draft against the high-budget distribution, retaining the accepted prefix
and sampling a continuation. Algorithm~\ref{alg:single_shot_self_speculative}
gives the full procedure.

Throughout, $\textsc{VerifyDraft}(\hat{a};\,q, r_d, r_v)$ denotes
standard speculative decoding verification, treating
$p_\theta(\cdot \mid q, r_d)$ as the proposal distribution and
$p_\theta(\cdot \mid q, r_v)$ as the target distribution, and
returning the length of the longest accepted prefix of $\hat{a}$.

\paragraph{Suffix decoding.}
After the accepted prefix is retained, generation continues
autoregressively under the verifier distribution. To accelerate
this continuation phase, we apply suffix decoding using a cache
built from the draft tokens $\hat{a}_1, \ldots, \hat{a}_m$. At
each step, the current generation context is used to look up
matching suffixes in the cache, proposing candidate spans that
are verified greedily against the verifier. Accepted spans
advance generation by multiple tokens per verifier step, reducing
the number of sequential decoding steps in the continuation phase.

This extension is most effective when the draft and final response
diverge in a minor early detail but agree on long spans later in the
text. This tends to hold in structured tasks such as coding and
planning, where the model commits to high-level structure early and
surface differences are localized.

\begin{algorithm}[t]
\caption{Single-Shot Self-Speculative Decoding from Partial Reasoning}
\label{alg:single_shot_self_speculative}
\begin{algorithmic}[1]
\Input Prompt $q$, draft budget $b_d$, verifier budget $b_v$, model $p_\theta$, suffix cache $C$
\Output Generated response $y$

\Statex \textbf{Phase 1: Concurrent CoT and Draft Generation}
\State $r \gets \emptyset$ \Comment{Reasoning trace}
\While{$|r| < b_v$}
    \State Sample $z \sim p_\theta(\cdot \mid q, r)$ \quad and \quad $r \gets \mathrm{concat}(r, z)$
    \If{$|r| = b_d$}
        \State $r_d \gets r$
        \State Start draft generation $\hat a \sim p_\theta(\cdot \mid q, r_d)$ in parallel \hfill $\parallel$
    \EndIf
    \If{draft generation finishes}
        \State Collect draft tokens $\hat a = (\hat a_1,\ldots,\hat a_m)$
        \quad and \quad $C \gets \mathrm{UpdateSuffixCache}(C, \hat a)$
    \EndIf
\EndWhile
\State $r_v \gets r$

\Statex
\Statex \textbf{Phase 2: Verification}
\State $k_{\mathrm{acc}} \gets \Call{VerifyDraft}{\hat a;\, q, r_d, r_v}$
\Statex \hspace{\algorithmicindent}\textit{$\hat a_{1:k_{\mathrm{acc}}}$ is the longest accepted draft prefix.} \Comment{draft CoT: $r_d$, verifier CoT: $r_v$}

\Statex
\Statex \textbf{Phase 3: Continuation}
\State Sample continuation $a_{\mathrm{cont}} \sim p_\theta(\cdot \mid q, r_v, \hat a_{1:k_{\mathrm{acc}}})$
\Statex \hspace{\algorithmicindent}\textit{Use suffix decoding with cache $C$ during continuation.}

\State \Return $\mathrm{concat}\left(r_v, \hat a_{1:k_{\mathrm{acc}}}, a_{\mathrm{cont}}\right)$
\end{algorithmic}
\end{algorithm}

\subsection{Serving-Time Implementation}

We implement self-speculation by duplicating the original request inside the scheduler when the draft budget is reached. The child request shares the same prefix as the original reasoning request, so vLLM's prefix caching can reuse the computation already performed for the prompt and partial CoT. The child then appends an end-of-thinking marker and begins answer drafting, while the parent request continues along the higher-budget reasoning path.

This implementation is simple and requires minimal changes to the serving stack, but it is not compute-free. During the overlap window, the scheduler is effectively serving both the parent and child requests, which can increase the instantaneous batch size and introduce bursts of additional compute. The speedup therefore comes from hiding draft latency behind ongoing reasoning, not from reducing total FLOPs. Prefix caching makes this practical by avoiding redundant computation over the shared prefix. We provide more details in the appendix.

\subsection{Multi-Stage and Iterative Variants}

The framework extends naturally to multi-stage decoding, where inference
budgets increase progressively. Given a sequence of budgets
$b_1 < b_2 < \cdots < b_T$, we apply speculative decoding iteratively:
at each stage, a draft generated under $b_i$ is verified and extended
under $b_{i+1}$. This yields a sequence of refinements in which
intermediate outputs are repeatedly reused rather than discarded.
Empirically, we observe that a substantial fraction of final-answer
tokens stabilize at relatively small budgets, which makes multi-stage
reuse effective in practice.

The single-stage version requires selecting a draft budget in advance
for sampling an initial response. This decision is important because if
the draft budget is too small, only a short prefix of the draft may be
accepted. Since the CoT length is not known in advance, this can require
domain-specific hyperparameter tuning. In the multi-stage version, the
draft is updated at regular intervals, making this choice less brittle:
one can safely choose an early initial draft budget and rely on later
stages to refine and extend the draft. The full procedure is given in
Algorithm~\ref{alg:multistage_self_speculative}  (Appendix~\ref{app:algorithms}).

\section{Experiments}
\label{experiments}

\begin{figure}[t]
\centering
\small
\setlength{\tabcolsep}{5pt}
\renewcommand{\arraystretch}{1.15}
\fbox{
\begin{minipage}{0.95\linewidth}
\centering

{\footnotesize\textbf{Schematic}}\\[0.2em]
\begin{tabular}{p{0.24\linewidth} p{0.66\linewidth}}
\toprule
\textbf{Mechanism} & \textbf{Accepted output structure} \\
\midrule
Prefix acceptance
&
\textcolor{PrefixAccepted}{\textbf{prefix prefix prefix}}
\;\;
\textcolor{RejectedDraft}{\textbf{rejected rejected}}
\\
Prefix + continuation
&
\textcolor{PrefixAccepted}{\textbf{prefix prefix prefix}}
\;\;
\textcolor{RegularDecoded}{cont cont}
\\
Prefix + suffix reuse
&
\textcolor{PrefixAccepted}{\textbf{prefix prefix prefix}}
\;\;
\textcolor{RegularDecoded}{cont cont}
\;\;
\textcolor{SuffixAccepted}{\textbf{suffix suffix}}
\;\;
\textcolor{RegularDecoded}{cont}
\\
\bottomrule
\end{tabular}

\vspace{0.35em}
{\centering
\rmfamily\footnotesize
\textcolor{PrefixAccepted}{\textbf{accepted prefix}};
\quad
\textcolor{RejectedDraft}{\textbf{rejected draft}};
\quad
\textcolor{RegularDecoded}{regular decoding};
\quad
\textcolor{SuffixAccepted}{\textbf{accepted suffix}}.
\par}

\vspace{0.55em}
\hrule
\vspace{0.45em}

{\footnotesize\textbf{Example}}\par
\vspace{0.2em}
\hrule
\vspace{0.45em}

\noindent
\begin{minipage}{0.94\linewidth}
\raggedright
\ttfamily\footnotesize
\begin{tabular}{@{}l@{}}
\textcolor{PrefixAccepted}{\textbf{from datetime import datetime, timedelta}} \\
\textcolor{PrefixAccepted}{\textbf{class CalendarUtil: ...}} \\
\textcolor{PrefixAccepted}{\textbf{def remove\_event(self, event): ...}} \\
\textcolor{PrefixAccepted}{\textbf{>>> calendar.events = [\{\sq date\sq: datetime(2023, 1, 1, 0, 0),}} \\
\textcolor{PrefixAccepted}{\textbf{... \quad \sq start\_time\sq: datetime}}
\textcolor{RegularDecoded}{(}
\textcolor{SuffixAccepted}{\textbf{2023, 1, 1, 0, 0}}
\textcolor{RegularDecoded}{),} \\
\textcolor{RegularDecoded}{... \quad \sq end}
\textcolor{SuffixAccepted}{\textbf{\_time}}
\textcolor{RegularDecoded}{\sq: datetime(2023, 1, 1, }
\textcolor{SuffixAccepted}{\textbf{1, 0}}
\textcolor{RegularDecoded}{),} \\
\textcolor{RegularDecoded}{... \quad \sq description\sq: \sq New Year\sq\}]} \\
\end{tabular}
\end{minipage}

\end{minipage}
}
\vspace{0.5em}
\caption{
Top: schematic illustration of prefix verification and suffix reuse.
Bottom: compressed real code-generation example showing the same structure.
}
\label{fig:suffix_decoding_combined}
\end{figure}

\begin{table}[t]
\centering
\small
\caption{Main latency results across benchmarks and 4B-class models.
Improvement is relative latency reduction, computed from mean per-sample
speedup as $100 \cdot (1 - 1/\mathrm{speedup})$.}
\label{tab:main_results}
\begin{tabular}{llrrrrrr}
\toprule
Model & Benchmark & $n$ & Improvement & SSR (s) & Base (s) & Prefix (tok) & Suffix (tok) \\
\midrule
Qwen3.5-4B & LongProc 2K & 989 & \gaincell{$9.1\%$} & $68.9$
& $72.8$ & $97.2$ & $529.6$ \\
Qwen3.5-4B & ClassEval & 100 & \gaincell{$18.5\%$} & $72.8$
& $85.3$ & $275.1$ & $674.4$ \\
Qwen3.5-4B & HumanEval & 164 & \gaincell{$2.9\%$} & $25.2$
& $26.4$ & $56.8$ & $88.6$ \\
\midrule
Gemma-4-E4B-it & LongProc 2K & 989 & \gaincell{$7.1\%$} & $97.4$
& $103.1$ & - & $556.7$ \\
Gemma-4-E4B-it & ClassEval & 100 & \gaincell{$24.1\%$} & $59.9$
& $79.2$ & $238.9$ & $637.8$ \\
Gemma-4-E4B-it & HumanEval & 164 & \gaincell{$14.6\%$} & $35.5$
& $41.8$ & $90.1$ & $225.5$ \\
\bottomrule
\end{tabular}
\end{table}

\begin{table}[t]
\centering
\small
\caption{Ablation of prefix verification and suffix decoding on ClassEval
with Gemma-4-E4B-it. Speedup is relative to same-token replay baseline.}
\label{tab:prefix_suffix_ablation_classeval_gemma}
\begin{tabular}{llccrrrr}
\toprule
Model & Benchmark & Prefix & Suffix & $n$ & Speedup &
Prefix tok. & Suffix tok. \\
\midrule
Gemma-4-E4B-it & ClassEval & \checkmark & -- &
100 & \gaincell{$1.086$} & $235.7$ & $0.0$ \\
Gemma-4-E4B-it & ClassEval & -- & \checkmark &
100 & \gaincell{$1.318$} & $0.0$ & $867.8$ \\
Gemma-4-E4B-it & ClassEval & \checkmark & \checkmark &
100 & \gaincell{$1.318$} & $238.9$ & $637.8$ \\
\bottomrule
\end{tabular}
\end{table}

\begin{table}[t]
\centering
\small
\caption{CoT throughput with and without concurrent drafting. Throughput is
computed as aggregate CoT tokens divided by aggregate CoT time. Draft share is
the fraction of CoT time during which a draft request was active.}
\label{tab:cot_draft_overhead}
\begin{tabular}{llrrrrr}
\toprule
Model & Benchmark & $n$ & No draft & Draft active & Change & Draft share \\
\midrule
Qwen3.5-4B & ClassEval & 100 & 40.20 & 38.03 & $-5.4\%$ & $29.5\%$ \\
Qwen3.5-4B & HumanEval & 164 & 40.20 & 38.12 & $-5.2\%$ & $25.5\%$ \\
Gemma-4-E4B-it & ClassEval & 100 & 36.70 & 36.68 & $-0.0\%$ & $30.9\%$ \\
Gemma-4-E4B-it & HumanEval & 164 & 36.50 & 36.46 & $-0.1\%$ & $33.2\%$ \\
\bottomrule
\end{tabular}
\end{table}

\begin{table}[t]
  \centering
  \small
  \caption{Stagewise latency breakdown on code benchmarks. Drafting may overlap
  with CoT generation, so components do not necessarily sum to total latency.
  Suffix decoding is included in continuation time.}
  \label{tab:stagewise_breakdown}
  \begin{tabular}{llrrrrrr}
  \toprule
  Model & Benchmark & CoT s & Draft s & Verify s & Cont. s &
  SSR total & Base total \\
  \midrule
  Qwen3.5-4B & ClassEval & 38.8 & 11.7 & 0.24 & 33.8
  & 72.8 & 85.3 \\
  Qwen3.5-4B & HumanEval & 20.9 & 8.4 & 0.15 & 4.6
  & 25.2 & 26.4 \\
  Gemma-4-E4B-it & ClassEval & 37.1 & 11.8 & 0.08 & 22.7
  & 59.9 & 79.2 \\
  Gemma-4-E4B-it & HumanEval & 27.8 & 10.8 & 0.07 & 7.7
  & 35.5 & 41.8 \\
  \bottomrule
  \end{tabular}
  \end{table}

\begin{table}[t]
\centering
\small
\caption{Iterative SSR reduces draft-child wall time. Draft wall time is the summed wall-clock lifetime of draft child requests spawned during CoT generation.}
\label{tab:draft_child_overhead}
\begin{tabular}{llrrrr}
\toprule
Model & Benchmark & Setting & Multi-drafting & Iterative SSR & Reduction \\
\midrule
Gemma-4-E4B-it & ClassEval & $i{=}750,\ m{=}500$  & $18.58$s & $16.37$s & \gaincell{$11.9\%$} \\
Gemma-4-E4B-it & ClassEval & $i{=}500,\ m{=}250$  & $17.75$s & $17.16$s & \gaincell{$3.3\%$} \\
\bottomrule
\end{tabular}
\end{table}

\subsection{Experimental Setup}
\label{experimental_setup}

We evaluate SSR in an end-to-end generation setting. Each request
generates a chain-of-thought up to a maximum reasoning budget; during
reasoning, SSR launches an answer draft from a lower-budget prefix,
verifies it under the higher-budget context, and continues generation
from the accepted prefix, optionally using suffix decoding with the
draft as an additional suffix-cache source. All latency comparisons
are against naive autoregressive generation with the same model and
sampling settings.

\textbf{Benchmarks.} Our methods are aimed at longer-form generation tasks where both the reasoning trace and the answer are significant contributors to generation latency. Coding is a strong fit, since it typically requires nontrivial reasoning and produces answers of meaningful length. For coding, we evaluate on ClassEval~\citep{classeval} and HumanEval~\citep{humaneval}: ClassEval contains longer class-level programs while HumanEval contains shorter function-level completions. Another primary benchmark is LongProc~\citep{longproc}, which contains long structured generation tasks with substantial variation across subtasks. We focus on the \textit{countdown}, \textit{html\_to\_tsv}, \textit{path\_traversal}, \textit{pseudo\_to\_code}, and \textit{tom\_tracking} subtasks, evaluating primarily on the 2K variant, which targets long outputs of roughly 2K tokens.

\textbf{Models.} We evaluate two 4B-class reasoning models: \texttt{Qwen3.5-4B}~\citep{qwenteam2026qwen35omnitechnicalreport} and \texttt{Gemma-4-E4B-IT}~\citep{gemma4_2026}. Both are open-source, independently developed, and deployable on a single GPU, providing a well-rounded comparison across model families.

\textbf{Metrics.} We report end-to-end speedup (baseline wall-clock latency divided by SSR latency) along with absolute latency. We additionally report accepted prefix tokens and suffix-accepted tokens to measure reuse from prefix verification and suffix decoding respectively.

\textbf{Compute.} All latency experiments run on a single machine with 10 NVIDIA RTX A6000 GPUs, with one vLLM worker per GPU and batch size 1. Tensor parallelism is not used for the 4B models. The machine has two 24-core Intel Xeon Gold 6342 CPUs.

\subsection{Results}
\label{results}

Table~\ref{tab:main_results} summarizes results across LongProc 2K, ClassEval, and HumanEval. The primary configuration uses prefix verification with suffix decoding. SSR gives the strongest gains on ClassEval, where Gemma-4-E4B-it achieves a $24.1\%$ latency reduction and Qwen3.5-4B achieves $18.5\%$. LongProc 2K shows moderate but consistent gains across both models ($7.1\%$--$9.1\%$). HumanEval shows the smallest improvements ($2.9\%$--$14.6\%$), which is expected: SSR only accelerates answer generation, so when CoT generation dominates total latency --- as it does for HumanEval's relatively short completions --- the gains are limited regardless of draft quality.

\textbf{Stagewise latency breakdown.} Tables~\ref{tab:stagewise_breakdown}
and~\ref{tab:cot_draft_overhead} decompose SSR latency and quantify the
cost of concurrent drafting. Draft generation takes $8$--$12$ seconds but
runs concurrently with CoT generation, so it does not add directly to
end-to-end latency. The throughput cost is small: Gemma-4-E4B-it shows
virtually no degradation, while Qwen3.5-4B incurs a $5.4\%$ reduction during
the draft-active window --- but since drafting is active for only
$25$--$30\%$ of CoT time, the effective overhead is under $2\%$, easily
offset by accepted draft tokens. Verification is negligible ($0.07$--$0.24$
seconds). The continuation phase is the primary remaining contributor: on
HumanEval, CoT generation dominates relative to continuation ($20.9$
vs.\ $4.6$ seconds for Qwen3.5-4B), explaining the smaller gains there.

\textbf{Prefix and suffix ablation.}
Table~\ref{tab:prefix_suffix_ablation_classeval_gemma} ablates the two reuse mechanisms on ClassEval with Gemma-4-E4B-it. Prefix-only verification yields a modest $7.9\%$ latency reduction, while suffix-only achieves $24.1\%$ despite accepting zero prefix tokens --- tokens that would have been accepted as a prefix are often recoverable as suffix matches, since they constitute exact lexical overlaps with the final answer.

\textbf{Iterative SSR overhead.} Table~\ref{tab:draft_child_overhead}
measures the wall-clock lifetime of draft child requests under multi-drafting
and iterative SSR, where $i$ is the draft interval and $m$ is the maximum
draft length. Iterative SSR reduces draft child wall time by $11.9\%$
at the $(i{=}750, m{=}500)$ setting and $3.3\%$ at $(i{=}500, m{=}250)$.
The larger reduction at the longer interval reflects that later drafts are
more likely to overlap with the final answer, so the bootstrapped draft
requires fewer tokens to complete.

\subsection{Discussion}
\label{discussion}

The preliminary results support three trends. First, SSR is most
effective when the final answer is long or structured enough for
accepted draft tokens to offset the overhead of drafting and
verification, explaining why LongProc 2K and ClassEval show
stronger speedups than HumanEval. Second, suffix decoding is
important because prefix verification alone discards useful draft
content after the first mismatch, especially in long-form outputs
with localized differences. Third, the method is not uniformly
beneficial: when drafts have low overlap with the final answer,
or when answer continuations are too short, the additional compute
can outweigh the latency hidden by concurrency.

\section{Limitations and Future Work}
\label{limitations_and_future_work}

SSR is most effective when answer length is comparable to CoT length;
when reasoning dominates generation time, gains are limited since CoT
generation is not accelerated. In high-entropy tasks, drafts may be
semantically similar to the final response but share little lexical
overlap, limiting prefix and suffix reuse. Natural directions for
future work include extending SSR to accelerate CoT generation itself
and fine-tuning models to commit to response structure earlier in the
reasoning trace.

\section{Conclusion}
\label{conclusion}

We introduce SSR, a training-free self-speculative decoding method for
reasoning language models that uses the chain-of-thought as a source of
speculation. By conditioning the same model at two reasoning budgets, SSR
derives both draft and verifier without auxiliary models or target-model-specific
modifications, hiding draft generation overhead behind concurrent reasoning.
Combined with suffix decoding to recover useful draft content beyond the
accepted prefix, SSR achieves consistent latency reductions on structured
and long-form generation tasks while preserving output quality. We hope
this work encourages exploration of reasoning structure as a
resource for efficient generation.

\nocite{*}
\bibliographystyle{plainnat}
\bibliography{references}

\appendix

\section{Serving-Time Implementation}
\label{app:serving-time-implementation}

When the reasoning request reaches the draft budget, the scheduler spawns
a child \textsc{Drafting} request initialized with the same token prefix,
followed by the model-specific end-of-thinking tokens. The child generates
up to a configured draft-token budget while the parent continues along the
longer reasoning path.

In multi-drafting mode, the scheduler spawns draft children at fixed token
increments, storing outputs by spawn point. When the parent finishes,
outstanding draft children are removed and the latest usable draft is
selected for verification.

Verification is implemented as a \textsc{Verification} request whose prompt
is the final reasoning prefix followed by the selected draft tokens. Prompt
log probabilities are compared against the draft's log probabilities and a
prefix is accepted using the standard speculative decoding rule. A
\textsc{Continuation} request then generates remaining answer tokens from
the accepted prefix, optionally using the draft as a suffix cache source.
All stages are ordinary vLLM requests coordinated by scheduler metadata,
with the final output returned under the original request id.

\section{Inference Hyperparameters}
\label{app:inference_hyperparameters}

Unless otherwise stated, all latency experiments use batch size 1 and run one
independent vLLM worker per GPU. Exploratory runs use eager mode, no warmup,
and one measured run. For each accelerated request, we run a separate
non-speculative vLLM instance as the baseline and generate the same number of
total response tokens as the accelerated request.

\begin{table}[h]
\centering
\small
\caption{Default inference and self-speculative decoding hyperparameters.}
\label{tab:inference_hyperparameters}
\begin{tabular}{ll}
\toprule
Hyperparameter & Value \\
\midrule
Precision & bfloat16 \\
Batch size & 1 \\
Tensor parallel size & 1 \\
GPU memory utilization & 0.9 \\
Runtime mode & eager \\
Warmup / measured runs & 0 / 1 \\
Sampling temperature & 0.6 \\
Top-$p$ / top-$k$ / min-$p$ & 1.0 / 0 / 0.0 \\
Self-spec verifier & rejection verification \\
Draft budget $b_d$ & 500 CoT tokens \\
Max draft length & 500 tokens \\
CoT budget $b_v$ & 2000 tokens \\
Suffix source & draft only \\
Internal suffix cache & disabled \\
Suffix tree depth & 4 \\
Speculative tokens & 16 \\
Suffix max spec factor & 4 \\
Suffix min token probability & 0.0 \\
Suffix max cached requests & 0 \\
\bottomrule
\end{tabular}
\end{table}

For Qwen-style thinking models, the end-of-thinking delimiter is
\texttt{</think>}. For Gemma-4-E4B-it, we use the native thinking boundary
\texttt{<channel|>}. When the model naturally emits the thinking boundary before the CoT cap, we truncate the CoT at that boundary and begin verification and
continuation from the resulting prefix. Continuation generation is allowed
to terminate at EOS before reaching the cap.

\section{Multi-Stage Self-Speculative Decoding}
\label{app:algorithms}

\begin{algorithm}[t]
\caption{Multi-Stage Self-Speculative Decoding from Partial Reasoning}
\label{alg:multistage_self_speculative}
\begin{algorithmic}[1]
\Input Prompt $q$, budgets $b_1 < b_2 < \cdots < b_T$, model $p_\theta$, suffix cache $C$
\Output Generated response $y$, updated suffix cache $C$

\State Generate reasoning up to budget $b_1$ to obtain $r_1$
\State Start draft generation $\hat a_1 \sim p_\theta(\cdot \mid q, r_1)$ in parallel

\For{$i = 1,\ldots,T-1$}
    \State Continue reasoning to budget $b_{i+1}$ to obtain $r_{i+1}$
    \State Collect draft $\hat a_i$ and update suffix cache $C$
    \State $k_i \gets \Call{VerifyDraft}{\hat a_i;\, q, r_i, r_{i+1}}$
    \Statex \hspace{\algorithmicindent}\textit{$\hat a_{i,1:k_i}$ is the longest prefix accepted under the higher-budget context $r_{i+1}$.}

    \If{$i < T-1$}
        \State Start next-stage draft generation
        \[
        \hat a_{i+1} \sim p_\theta(\cdot \mid q, r_{i+1}, \hat a_{i,1:k_i})
        \]
        \Statex \hspace{\algorithmicindent}\textit{Reuse the accepted prefix rather than drafting from scratch.}
    \EndIf
\EndFor

\State Sample continuation $a_{\mathrm{cont}} \sim p_\theta(\cdot \mid q, r_T, \hat a_{T-1,1:k_{T-1}})$
\Statex \hspace{\algorithmicindent}\textit{Use suffix decoding with cache $C$ during continuation.}

\State \Return $\mathrm{concat}\left(r_T, \hat a_{T-1,1:k_{T-1}}, a_{\mathrm{cont}}\right), C$

\end{algorithmic}
\end{algorithm}

\newpage
\section*{NeurIPS Paper Checklist}

The checklist is designed to encourage best practices for responsible machine learning research, addressing issues of reproducibility, transparency, research ethics, and societal impact. Do not remove the checklist: {\bf The papers not including the checklist will be desk rejected.} The checklist should follow the references and follow the (optional) supplemental material.  The checklist does NOT count towards the page
limit. 

Please read the checklist guidelines carefully for information on how to answer these questions. For each question in the checklist:
\begin{itemize}
    \item You should answer \answerYes{}, \answerNo{}, or \answerNA{}.
    \item \answerNA{} means either that the question is Not Applicable for that particular paper or the relevant information is Not Available.
    \item Please provide a short (1--2 sentence) justification right after your answer (even for \answerNA). 
\end{itemize}

{\bf The checklist answers are an integral part of your paper submission.} They are visible to the reviewers, area chairs, senior area chairs, and ethics reviewers. You will also be asked to include it (after eventual revisions) with the final version of your paper, and its final version will be published with the paper.

The reviewers of your paper will be asked to use the checklist as one of the factors in their evaluation. While \answerYes{} is generally preferable to \answerNo{}, it is perfectly acceptable to answer \answerNo{} provided a proper justification is given (e.g., error bars are not reported because it would be too computationally expensive'' or ``we were unable to find the license for the dataset we used''). In general, answering \answerNo{} or \answerNA{} is not grounds for rejection. While the questions are phrased in a binary way, we acknowledge that the true answer is often more nuanced, so please just use your best judgment and write a justification to elaborate. All supporting evidence can appear either in the main paper or the supplemental material, provided in appendix. If you answer \answerYes{} to a question, in the justification please point to the section(s) where related material for the question can be found.

IMPORTANT, please:
\begin{itemize}
    \item {\bf Delete this instruction block, but keep the section heading ``NeurIPS Paper Checklist"},
    \item  {\bf Keep the checklist subsection headings, questions/answers and guidelines below.}
    \item {\bf Do not modify the questions and only use the provided macros for your answers}.
\end{itemize}


\begin{enumerate}

\item {\bf Claims}
    \item[] Question: Do the main claims made in the abstract and introduction accurately reflect the paper's contributions and scope?
    \item[] Answer: \answerYes
    \item[] Justification: We describe the settings and scope intended for the outlined methods within the paper.

\item {\bf Limitations}
    \item[] Question: Does the paper discuss the limitations of the work performed by the authors?
    \item[] Answer: \answerYes.
    \item[] Justification: We add a limitations section to describe settings in which our algorithm will likely struggle.

\item {\bf Theory assumptions and proofs}
    \item[] Question: For each theoretical result, does the paper provide the full set of assumptions and a complete (and correct) proof?
    \item[] Answer: \answerNA{}.
    \item[] Justification: No theoretical results states in this paper.

    \item {\bf Experimental result reproducibility}
    \item[] Question: Does the paper fully disclose all the information needed to reproduce the main experimental results of the paper to the extent that it affects the main claims and/or conclusions of the paper (regardless of whether the code and data are provided or not)?
    \item[] Answer: \answerYes.
    \item[] Justification: We outline the datasets and inference settings in the experiments section, and plan on releasing code upon publication.

\item {\bf Open access to data and code}
    \item[] Question: Does the paper provide open access to the data and code, with sufficient instructions to faithfully reproduce the main experimental results, as described in supplemental material?
    \item[] Answer: \answerNo
    \item[] Justification: We intend to release our code upon publication.

\item {\bf Experimental setting/details}
    \item[] Question: Does the paper specify all the training and test details (e.g., data splits, hyperparameters, how they were chosen, type of optimizer) necessary to understand the results?
    \item[] Answer: \answerYes 
    \item[] Justification: Described in the experiments section, and within the evaluation scripts.

\item {\bf Experiment statistical significance}
    \item[] Question: Does the paper report error bars suitably and correctly defined or other appropriate information about the statistical significance of the experiments?
    \item[] Answer: \answerNo{}
    \item[] Justification: As in other similar works, we report only the results obtained by standard evaluation scripts, which might not include statistical signficance.

\item {\bf Experiments compute resources}
    \item[] Question: For each experiment, does the paper provide sufficient information on the computer resources (type of compute workers, memory, time of execution) needed to reproduce the experiments?
    \item[] Answer: \answerYes.
    \item[] Justification: We mention the resources used within the experiments section.
    
\item {\bf Code of ethics}
    \item[] Question: Does the research conducted in the paper conform, in every respect, with the NeurIPS Code of Ethics \url{https://neurips.cc/public/EthicsGuidelines}?
    \item[] Answer: \answerYes.
    \item[] Justification: We believe we have abided by the NeurIPS code of ethics in the submission of this paper.

\item {\bf Broader impacts}
    \item[] Question: Does the paper discuss both potential positive societal impacts and negative societal impacts of the work performed?
    \item[] Answer: \answerNA{}.
    \item[] Justification: This work will likely have no direct societal impact.
    
\item {\bf Safeguards}
    \item[] Question: Does the paper describe safeguards that have been put in place for responsible release of data or models that have a high risk for misuse (e.g., pre-trained language models, image generators, or scraped datasets)?
    \item[] Answer: \answerNA{}
    \item[] Justification: This work has low risk of misuse, as it is an inference-time optimization.

\item {\bf Licenses for existing assets}
    \item[] Question: Are the creators or original owners of assets (e.g., code, data, models), used in the paper, properly credited and are the license and terms of use explicitly mentioned and properly respected?
    \item[] Answer: \answerYes 
    \item[] Justification: We believe we have credited the original owners of the main code/data/models used in this work. 

\item {\bf New assets}
    \item[] Question: Are new assets introduced in the paper well documented and is the documentation provided alongside the assets?
    \item[] Answer: \answerYes
    \item[] Justification: Our code is accompanied with relevant documentation.

\item {\bf Crowdsourcing and research with human subjects}
    \item[] Question: For crowdsourcing experiments and research with human subjects, does the paper include the full text of instructions given to participants and screenshots, if applicable, as well as details about compensation (if any)? 
    \item[] Answer: \answerNA{} 
    \item[] Justification: No human subjects or sources used for this paper.

\item {\bf Institutional review board (IRB) approvals or equivalent for research with human subjects}
    \item[] Question: Does the paper describe potential risks incurred by study participants, whether such risks were disclosed to the subjects, and whether Institutional Review Board (IRB) approvals (or an equivalent approval/review based on the requirements of your country or institution) were obtained?
    \item[] Answer: \answerNA{} 
    \item[] Justification: No crowdsourcing with human subjects.

\item {\bf Declaration of LLM usage}
    \item[] Question: Does the paper describe the usage of LLMs if it is an important, original, or non-standard component of the core methods in this research? Note that if the LLM is used only for writing, editing, or formatting purposes and does \emph{not} impact the core methodology, scientific rigor, or originality of the research, declaration is not required.
    \item[] Answer: \answerNA{} 
    \item[] Justification: LLMs used only for writing/editing/formatting.

\end{enumerate}

\end{document}